\documentclass{article}

\usepackage[final]{corl_2026}    

\usepackage{graphicx}
\usepackage[table]{xcolor}
\usepackage{booktabs}
\usepackage{tabularx}
\usepackage{array}
\usepackage{amsmath}
\usepackage{hyperref}
\usepackage{amssymb}
\usepackage{multirow}
\usepackage{makecell}
\usepackage{subcaption}
\usepackage{csquotes}

\title{When, Where, and What to Correct: Human-Guided Adaptive Residual Correction for Robot Policies}

\author{
\textbf{Anukriti Singh$^{*}$} \\
University of Maryland, College Park \\
College Park, MD \\
\texttt{anukriti@umd.edu}
\And
\textbf{Maxime Caniot} \\
Enchanted Tools \\
Paris, France
\And
\textbf{Pratap Tokekar} \\
University of Maryland, College Park \\
College Park, MD
\And
\textbf{Firas Abi-Farraj} \\
Enchanted Tools \\
Paris, France
}

\begin{document}
\maketitle

\begingroup
\renewcommand{\thefootnote}{*}
\footnotetext[1]{Work done during internship at Enchanted Tools.}
\endgroup

\begin{abstract}
Imitation policies that reliably execute most of a manipulation trajectory can
still fail in localized regions requiring last-mile precision. Updating the full
policy or exploring residual actions everywhere treats this local failure as a
global learning problem. We introduce \textbf{Adaptive Residual Correction
(ARC)}, a lightweight post-training module that uses reinforcement learning
(RL) to learn a residual correction on top of a frozen policy while using sparse
human corrections to structure autonomous exploration. Rather than treating
human corrections as imitation targets, ARC uses pre-intervention states to
determine \emph{when} residual control should become active and correction
geometry to determine \emph{where in residual action space} autonomous
refinement should explore; task reward determines \emph{what} residual is
ultimately executed. Across three precision-manipulation tasks on a real robot,
ARC improves average task success from $33.3\%$ to $75.0\%$. The human
corrections define task-dependent 2--3-dimensional subspaces of the seven-joint
residual space that retain $96.0$--$97.7\%$ of squared correction magnitude.
Together, these results show that sparse human corrections can localize and
constrain autonomous post-training without requiring the observed correction
itself to serve as the final action target
(\href{https://anonymous.4open.science/w/arc-6DD5/}{link}).
\end{abstract}


\section{Introduction}
\label{sec:introduction}

Modern imitation-learning policies can acquire broad manipulation competence
from demonstrations, including complex closed-loop visuomotor behaviors
\cite{zhao2023learningfinegrainedbimanualmanipulation,chi2024diffusionpolicyvisuomotorpolicy}.
Yet even a policy that reliably reaches the relevant interaction can remain
brittle to small execution errors and distribution shift, particularly when
success depends on a short precision-critical segment of the task
\cite{zhao2023learningfinegrainedbimanualmanipulation,ankile2025resip}. In this
regime, collecting substantially more demonstrations or post-training the
entire policy treats a localized failure as a global learning problem. Residual
learning instead keeps the base policy fixed and learns only the additional
control required to correct its remaining failures
\cite{ankile2025resip,yuan2025policydecorator,
ankile2025residualoffpolicyrlfinetuning,kowshik2026midas}.

Residual learning is most useful when the correction itself is similarly
targeted. Acting throughout the trajectory can degrade behavior the base policy
already performs well, while arbitrary action-space exploration wastes
interaction on unlikely motions. Sparse human corrections provide structure
for both problems. During an otherwise autonomous rollout, the human intervenes
only when needed by adding a residual adjustment to the base-policy action; we
refer to these intermittent adjustments as \emph{human corrections}.
Human-in-the-loop methods show that sparse interventions can efficiently guide
policy improvement \cite{luo2025hilserl,zhao2025silri}, while intervention
timing itself carries information beyond the corrective trajectory
\cite{wang2026correctiontiming}. At autonomous deployment, no human is present:
the residual policy must recognize robot-controlled states in which a human
\emph{would soon have corrected}, rather than states already altered by a
correction. Likewise, the coordinated multi-joint geometry of the correction
reveals which residual directions the human actually uses. Together, these
signals localize the temporal and action-space support in which residual
learning should operate.

\begin{figure}[t]
    \centering
    \includegraphics[width=\linewidth]{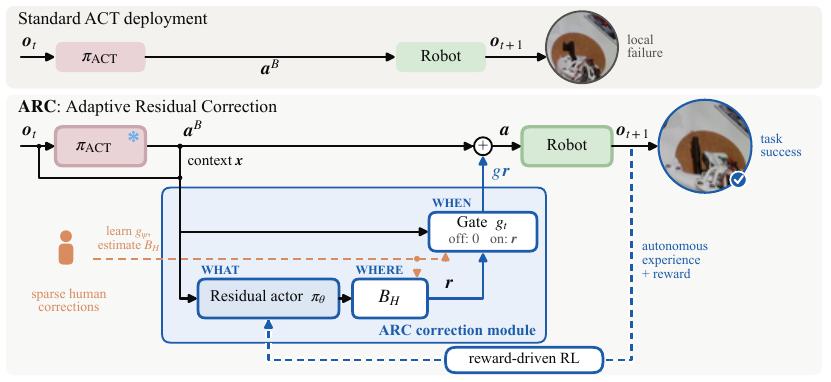}
    \caption{\textbf{ARC overview.} Sparse human corrections train a
    pre-intervention gate and define a low-dimensional corrective subspace,
    which together constrain autonomous exploration in time and action space.
    Task reward trains the residual actor, whose gated correction is
    added to the frozen ACT action at deployment.}
    \label{fig:arc_overview}
\end{figure}


The human correction itself, however, need not be the final action an
autonomous policy should reproduce. Recent methods use corrective actions
directly, or through latent representations, as supervision for learned
steering policies \cite{murray2026flowdagger,lu2026unisteer}, but a correction
is chosen to recover from the particular state encountered during intervention,
and directly imitating it does not encode whether its downstream consequence
leads to task success \cite{zhao2025silri}. For example, a locally sensible
placement correction may still disturb another object and cause the task to
fail.

This distinction motivates \textbf{Adaptive Residual Correction (ARC)}, a
lightweight plug-in module for improving a frozen base policy. During
otherwise autonomous rollouts, the human provides sparse residual corrections
only when needed. From approximately 8--10 such rollouts per task, ARC learns a
pre-intervention gate identifying \textbf{WHEN} residual control should become
active and a low-dimensional basis of the seven-joint residual action space
restricting \textbf{WHERE} autonomous refinement searches. These correction
rollouts are used only to learn the gate and corrective basis and are not added
to the RL replay buffer. ARC then collects separate autonomous experience
within this learned temporal and action-space support, and reinforcement
learning determines \textbf{WHAT} correction improves task outcome. The base
policy remains frozen and its original demonstrations are not revisited.

Relative to residual-RL methods
\cite{ankile2025resip,yuan2025policydecorator,
ankile2025residualoffpolicyrlfinetuning,kowshik2026midas}, ARC restricts both
when the residual is active and where exploration occurs; relative to
correction-supervised methods \cite{murray2026flowdagger,lu2026unisteer}, it
uses human residuals to define the support of autonomous learning rather than
as targets for the final residual actor.

We evaluate ARC on three precision-manipulation tasks on a real robot:
aligning and stacking industrial trays, inserting a marker into a narrow groove,
and stacking cups into a fixed receiver. After two autonomous
collect--reward--update rounds, ARC improves average task success from $33.3\%$
to $75.0\%$. Experiments further show that pre-intervention states yield a gate
that transfers to autonomous execution, human residuals define stable
low-dimensional corrective supports, and directly predicting the human
correction produces residual behavior different from reward-driven ARC even
when supervision is restricted to the same corrective subspace.


\vspace{-4mm}
\section{Related Work}
\label{sec:related}
\vspace{-4mm}
\textbf{Residual refinement.}
ResiP~\cite{ankile2025resip} adds closed-loop residual RL to a frozen
action-chunked policy, while Policy Decorator~\cite{yuan2025policydecorator},
ResFiT~\cite{ankile2025residualoffpolicyrlfinetuning}, and
MiDAS~\cite{kowshik2026midas} study residual post-training and low-data
adaptation. These methods do not by themselves determine when refinement
should be active or how human corrections should constrain residual search.

\textbf{Learning from corrections.}
HIL-SERL~\cite{luo2025hilserl} incorporates online interventions into
real-world RL, SiLRI~\cite{zhao2025silri} balances imitation and reinforcement
from potentially suboptimal interventions, and
FlowDAgger~\cite{murray2026flowdagger} and UniSteer~\cite{lu2026unisteer}
transform corrections into supervision for steering policies. ARC instead
uses corrections primarily to define temporal and action-space support rather
than as targets for its reward-trained actor.

\textbf{Structuring refinement.}
DGN~\cite{dong2025dgn} shapes exploration from demonstrations,
SART~\cite{oh2025sart} restricts variation to human-identified precision
regions, Wang et al.~\cite{wang2026correctiontiming} study intervention timing,
and FRS~\cite{tang2026frs} uses coarse guidance to bootstrap RL. ARC uses the
same sparse correction to define both temporal and action-space support, while
task reward determines the final residual.

\section{Adaptive Residual Correction (ARC)}
\label{sec:method}
\vspace{-2mm}
\subsection{Problem Formulation}
\label{sec:problem}
\vspace{-2mm}

Let $o_t\in\mathcal{O}$ denote the robot observation and
$a_t\in\mathcal{A}\subseteq\mathbb{R}^7$ the seven-joint arm action at time
$t$; the gripper remains controlled by the base policy. We study post-training
of a pretrained policy $\pi_B$, with $a_t^B=\pi_B(o_t)$, using sparse human
corrections while keeping $\pi_B$ frozen. During otherwise autonomous rollouts,
a human uses a SpaceMouse, a six-degree-of-freedom handheld controller, to add a
residual adjustment to the base action rather than replacing it. If $a_t^H$ is
the resulting action, the human residual is $r_t^H=a_t^H-a_t^B$.

A correction contains information beyond this instantaneous action: its onset
indicates when the base trajectory is becoming unreliable, while its
coordinated joint residual identifies useful corrective directions. ARC uses
both signals to restrict autonomous learning instead of treating $r_t^H$ as
the final policy target. We write the adapted action as
$a_t=a_t^B+g_t r_t$, where $g_t\in\{0,1\}$ is a binary gate and
$r_t=B_Hz_t$ is the residual action. Here,
$B_H\in\mathbb{R}^{7\times d_H}$ is a human-derived corrective basis and
$z_t\in\mathbb{R}^{d_H}$ is its coefficient vector. Thus, $B_Hz_t$ maps the
low-dimensional coefficients back to the seven-joint residual action space.
Residual components operate on $x_t=[\phi_B(I_t),q_t,a_t^B]$, where $I_t$ is
the image observation, $\phi_B(I_t)$ is the frozen ACT visual representation,
and $q_t$ is proprioception.

%

\subsection{WHEN: Localizing Residual Refinement}
\label{sec:when}
\vspace{-2mm}

A classifier trained on states during an active human correction may rely on
states already altered by the human. ARC instead predicts an upcoming
correction from robot-controlled states. Using only the first state of each
correction bout would provide one positive example per bout; we instead use a
short preceding window to provide more positive examples and represent a small
region where correction is imminent rather than one exact frame. For each
correction bout beginning at $t_0$, the preceding $L=5$ states are positive
examples; correction-active frames are excluded and sufficiently distant
robot-controlled states form negatives.

We train a binary gate classifier $g_\psi$ that outputs
$p_t=g_\psi(x_t)=P(y_t=1\mid x_t)$, where $y_t=1$ denotes a correction
beginning within the next $L$ states. Episode-level splits prevent nearby
frames from the same rollout appearing in both training and validation. At
deployment, $p_t$ is thresholded and passed through temporal hysteresis to
produce the binary gate $g_t$ used in $a_t=a_t^B+g_t r_t$. The probability
threshold is chosen from leave-one-episode-out predictions using Youden's $J$
statistic, while the on/off durations are fixed from human correction timing
statistics. These values are not tuned using autonomous deployment rollouts.
When $g_t=0$, ARC exactly recovers the frozen base action.

\subsection{WHERE: Human-Supported Corrective Action Space}
\label{sec:where}
\vspace{-2mm}
Human corrections induce coordinated changes across joints, motivating a
low-dimensional corrective basis rather than independent joint perturbations.
Because zero residual has a meaningful physical interpretation, for residuals
$\{r_i^H\}_{i=1}^{N}$ we compute the uncentered second moment
$M_H=\frac{1}{N}\sum_{i=1}^{N}r_i^H(r_i^H)^\top$, whose leading eigenvectors
form the corrective basis $B_H=[v_1,\ldots,v_{d_H}]$. We select candidate
subspaces using the fraction of total squared residual magnitude captured by
the retained directions together with leave-one-episode-out stability. Once
selected, $B_H$ is frozen for autonomous interaction, and we estimate a human
correction scale $\sigma_k$ from the projected correction magnitudes along each
retained direction.

\vspace{-3mm}
\subsection{WHAT: Reward-Driven Residual Learning}
\label{sec:what}
\vspace{-2mm}
\paragraph{Human-guided exploration.}
The gate and corrective basis jointly define autonomous exploration. At the
start of each rollout, we choose one retained corrective direction $v_k$
(a column of $B_H$), with $k\in\{1,\ldots,d_H\}$, and sign
$s\in\{-1,+1\}$. We use the fixed episode-level residual
$\epsilon=s\,\alpha\,\sigma_kv_k$, giving
$a_t=a_t^B+g_t\epsilon$, where $\alpha$ controls exploration magnitude.
Holding the signed direction fixed provides coherent action--outcome variation,
while the gate prevents perturbations outside the selected refinement region.
After controller limits are applied, let $a_t^{\mathrm{exec}}$ denote the
action physically executed by the robot. Replay stores the corresponding
residual $r_t^{\mathrm{exec}}=a_t^{\mathrm{exec}}-a_t^B$, so the critic learns
from actions that actually occurred on the robot. Human correction trajectories
themselves are not inserted into replay.

%
%
%

\textbf{Residual learning.}
Once WHEN and WHERE are fixed, ARC learns the correction from autonomous task
reward rather than treating the human residual as an action target. The actor
predicts normalized coefficients
$u_t=\pi_\theta(x_t)\in[-1,1]^{d_H}$. These are scaled by the human-derived
correction magnitudes to obtain the coefficient-space residual action
$z_t=(\alpha\sigma_H)\odot u_t$, where
$\sigma_H=[\sigma_1,\ldots,\sigma_{d_H}]^\top$. The resulting joint-space
residual is
\begin{equation}
    r_\theta(x_t)
    =
    B_Hz_t
    =
    B_H\left[(\alpha\sigma_H)\odot\pi_\theta(x_t)\right].
    \label{eq:actor_residual}
\end{equation}
We optimize this residual using TD3
\cite{fujimoto2018addressingfunctionapproximationerror}, multi-step TD targets,
and a regularizer toward small residuals. Human actions are not regression
targets in this stage: they define the support of refinement, while autonomous
reward determines the residual within that support. At deployment, exploration
is removed and the learned residual is gated as
$a_t=a_t^B+g_t r_\theta(x_t)$.
\vspace{-4mm}
%
%
%

\begin{figure}[!t]
    \centering
    \includegraphics[width=\linewidth]{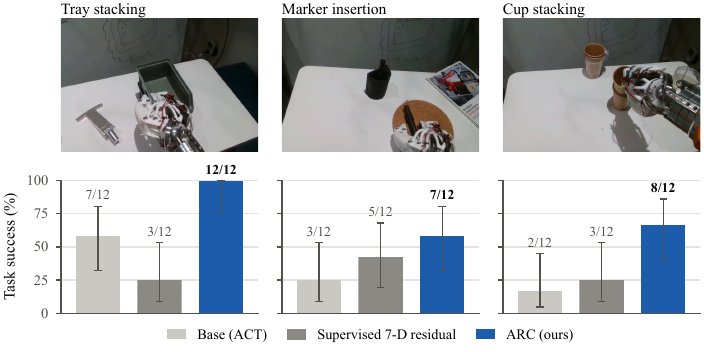}
    \caption{\textbf{Performance on a real robot.}
    Base ACT, direct supervised seven-joint residual correction, and ARC over
    12 trials per method and task. Labels show exact successes; error bars are
    $95\%$ Wilson confidence intervals.}
    \label{fig:overall_results}
\end{figure}

\section{Experimental Evaluation}
\label{sec:evaluation}
\vspace{-2mm}

We ask:
\textbf{Q1} whether ARC improves last-mile precision;
\textbf{Q2} what sparse corrections reveal about \emph{when} and \emph{where}
to refine; and
\textbf{Q3} whether the observed correction itself should determine
\emph{what} the robot executes.

\subsection{Experimental Setup}
\label{sec:setup}

We evaluate ARC on the Miroki humanoid robot using its right arm and a custom
3D-printed claw gripper. Each task uses a frozen pretrained ACT policy. The
three tasks are tray stacking, marker insertion into a narrow groove, and cup
stacking into a fixed receiver (Fig.~\ref{fig:overall_results}). The base
policies are trained from approximately 55 demonstrations per task on average.
For post-training, we collect approximately 8--10 base-policy rollouts per task
in which the human intermittently adds residual corrections. These
correction-containing rollouts provide the pre-intervention states used to
train ARC's gate and the human residuals used to construct its corrective
basis; human-corrected transitions are not inserted into the RL replay buffer.
ARC then performs two autonomous collect--reward--update rounds. We compare
Base ACT, a supervised baseline trained on the same correction data to directly
regress the observed seven-joint human residuals, and ARC over 12 real-robot
trials per method and task. We report exact successes and $95\%$ Wilson
confidence intervals.

%
%

\begin{figure}[!t]
    \centering
    \includegraphics[width=0.75\linewidth]{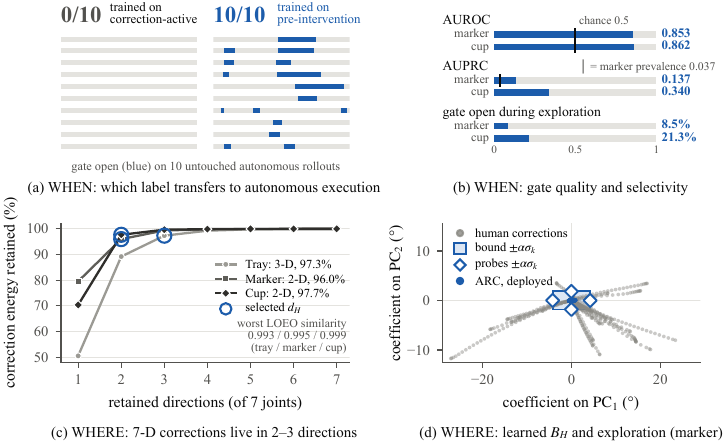}
    \caption{\textbf{WHEN and WHERE.}
    (a) On 10 untouched autonomous base-policy rollouts, a gate trained on
    correction-active states never activates, while the pre-intervention gate
    activates on all 10 (gate activation, not task success).
    (b) Held-out gate discrimination and the fraction of autonomous-exploration
    timesteps for which the gate is open.
    (c) The selected $3$-, $2$-, and $2$-D subspaces retain $97.3\%$,
    $96.0\%$, and $97.7\%$ of squared correction magnitude for tray, marker,
    and cup, respectively.
    (d) Marker corrections, exploration probes, and deployed ARC residuals in
    the learned corrective basis; PC denotes principal component.}
    \label{fig:when_where_results}
\end{figure}

\vspace{-2mm}
\subsection{Q1: Can ARC Improve Last-Mile Precision?}
\label{sec:overall_results}
\vspace{-2mm}
As shown in Fig.~\ref{fig:overall_results}, ARC improves the frozen base policy
from $58.3\%$ ($7/12$) to $100.0\%$ ($12/12$) on tray stacking, from
$25.0\%$ ($3/12$) to $58.3\%$ ($7/12$) on marker insertion, and from
$16.7\%$ ($2/12$) to $66.7\%$ ($8/12$) on cup stacking. Across all three
tasks, the base policy reaches the relevant manipulation phase, but small
execution errors become consequential during tray alignment, near the marker
groove, or while placing a cup without disturbing the receiver. ARC therefore
acts as a selective correction layer rather than a replacement policy. Tray
stacking is also the most precision-demanding task in our setup and exhibits
three corrective directions rather than two for the other tasks, consistent
with the lower performance of direct supervised correction
(Fig.~\ref{fig:when_where_results}(c)).

\vspace{-2mm}
\subsection{Q2: What Do Human Corrections Tell Us About WHEN and WHERE?}
\label{sec:when_where_results}
\vspace{-2mm}
\textbf{WHEN.}
As a baseline, we train a correction-active classifier using states from within
human correction bouts as positives. In contrast, the pre-intervention gate is
trained on the preceding robot-controlled states, matching the condition
encountered at autonomous deployment. This distinction matters: the former
activates on none of 10 independently collected autonomous base-policy
rollouts, while the latter activates on all 10
(Fig.~\ref{fig:when_where_results}(a)). For marker insertion, the
pre-intervention gate achieves an area under the receiver operating
characteristic curve (AUROC) of $0.853$ and an area under the precision--recall
curve (AUPRC) of $0.137$, compared with a positive-class prevalence of $0.037$,
and identifies $17/36$ held-out correction onsets. The cup gate achieves AUROC
$0.862$ and AUPRC $0.340$. The learned gate remains selective during autonomous
exploration: residual control is active for only $8.5\%$ of marker and
$21.3\%$ of cup timesteps.

\begin{figure}[!t]
    \centering
    \includegraphics[width=0.75\linewidth]{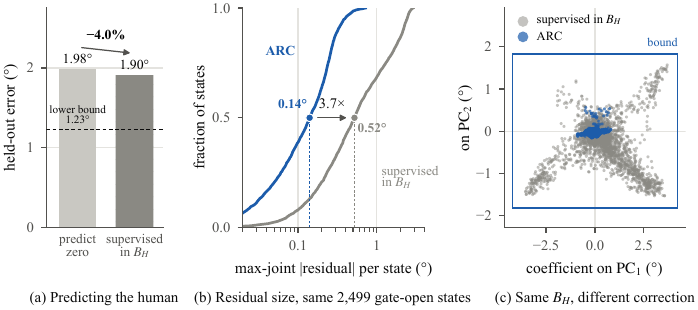}
    \caption{\textbf{WHAT: same support, different corrections.}
    (a) Within $B_H$, supervised prediction reduces held-out human-residual
    error only from $1.98^\circ$ to $1.90^\circ$ ($4.0\%$), while oracle
    projection gives a $1.23^\circ$ lower bound.
    (b) On the same 2,499 gate-open states, median max-joint residual magnitude
    is $0.52^\circ$ for supervision and $0.14^\circ$ for ARC ($3.7\times$).
    (c) With the same $B_H$ and residual bounds, PC$_1$ sign agreement is
    $38\%$ on non-trivial states, while $22\%$ of supervised versus $0\%$ of
    ARC predictions reach at least half the allowed bound.}
    \label{fig:what_results}
\end{figure}

\textbf{WHERE.}
Temporal localization still leaves a seven-joint residual space. Human
corrections instead concentrate in compact and reproducible subspaces
(Fig.~\ref{fig:when_where_results}(c--d)): tray stacking requires three
directions, while marker insertion and cup stacking require two, retaining
$97.3\%$, $96.0\%$, and $97.7\%$ of squared correction magnitude. Worst
leave-one-episode-out subspace similarities are $0.993$, $0.995$, and $0.999$.
For marker insertion, the same learned basis defines the bounded exploration
probes and contains the residuals produced by deployed ARC
\vspace{-3mm}

%
%
%
\subsection{Q3: Should Human Corrections Define WHAT to Execute?}
\label{sec:what_results}
\vspace{-2mm}

Our primary supervised baseline directly predicts the seven-joint human
residual $r_t^H$ from the same context $x_t$ and applies it through the same
learned gate. It achieves $3/12$, $5/12$, and $3/12$ successes on tray,
marker, and cup, compared with ARC's $12/12$, $7/12$, and $8/12$
(Fig.~\ref{fig:overall_results}). This comparison differs in both
\emph{WHAT}---direct regression versus reward learning---and
\emph{WHERE}---the full seven-joint residual space versus
$\operatorname{span}(B_H)$. To isolate WHAT, we therefore train an additional
WHERE-matched supervised policy using the same human corrections projected into
$B_H$ and the same residual bounds as ARC.

Even after matching WHERE, supervision does not recover ARC's residual
behavior. As Fig.~\ref{fig:what_results} shows, supervised prediction only
slightly improves held-out human-residual error and produces substantially
larger residuals than ARC on the same gate-open states. Panel (c) further shows
that the two policies select different coefficients within the same corrective
support. Thus, human corrections provide useful structure for refinement, but
directly imitating them does not reproduce the correction selected from
autonomous task reward.

%

\vspace{-3mm}
\section{Conclusion and Discussion}
\label{sec:conclusion}
\vspace{-3mm}

ARC uses sparse human corrections to identify \emph{when} residual control is
needed and \emph{where} in action space to refine, while autonomous task reward
determines \emph{what} residual is executed. Across three tasks on a real robot,
this targeted post-training improves last-mile precision without updating the
base policy or revisiting its original demonstrations.

\textbf{Limitations.}
ARC currently relies on human correction timing to train the gate that switches
residual control on and off. This requires collecting task- and
operator-specific correction examples, and the resulting gate may vary with
how different humans choose when to intervene. A promising direction is to
replace or augment this learned gate with a continuously operating
vision-language-model monitor that could recognize precision-critical states
and determine when residual refinement is needed, reducing reliance on
task-specific human correction timing.

\bibliography{example}

\end{document}